%% file: paper_for_arxiv.tex
\documentclass[10pt,twocolumn,letterpaper]{article}

\usepackage{cvpr}
\usepackage{times}
\usepackage{epsfig}
\usepackage{graphicx}
\usepackage{amsmath}
\usepackage{amssymb}
\usepackage{caption}
\usepackage{subcaption}
\usepackage[utf8]{inputenc}\usepackage[utf8]{inputenc}
\usepackage{balance}
\usepackage{comment}

\usepackage[pagebackref=true,breaklinks=true,letterpaper=true,colorlinks,bookmarks=false]{hyperref}

\cvprfinalcopy %

\def\cvprPaperID{6155} %

\DeclareMathOperator*{\argmin}{argmin}

\begin{document}

\input{macros}

\title{\OURS: Learning Non-rigid RGB-D Reconstruction \\ with Semi-supervised Data}

\author{
Aljaž Božič$^{1}$ \qquad Michael Zollhöfer$^{2}$  \qquad Christian Theobalt$^{3}$ \qquad Matthias Nie{\ss}ner$^{1}$ 
\vspace{0.2cm} \\ 
$^{1}$Technical University of Munich \qquad $^{2}$Stanford University \qquad $^{3}$Max Planck Institute for Informatics 
\vspace{0.2cm} \\
}

\newcommand\blfootnote[1]{%
  \begingroup
  \renewcommand\thefootnote{}\footnote{#1}%
  \addtocounter{footnote}{-1}%
  \endgroup
}

\twocolumn[{%
	\renewcommand\twocolumn[1][]{#1}%
	\maketitle
	\begin{center}
		\vspace{-0.8cm}
		\includegraphics[width=0.96\linewidth]{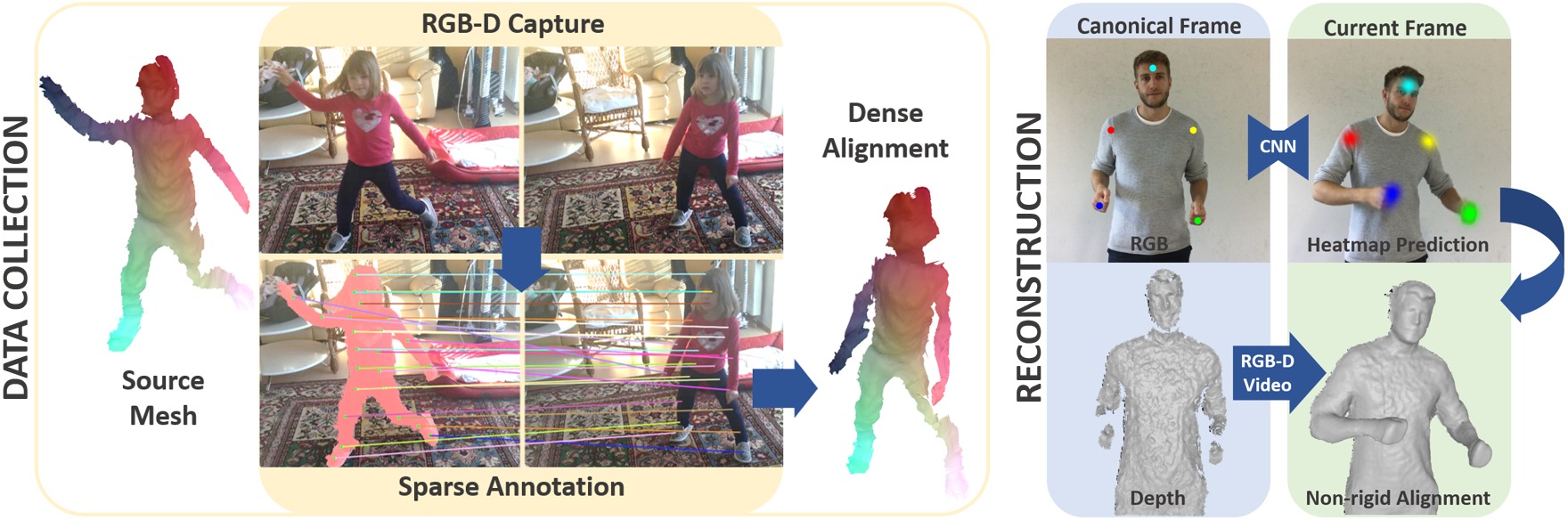}
		\vspace{-0.2cm}
		\captionof{figure}{
	We propose a semi-supervised strategy combining self-supervision with sparse annotations to build a large-scale RGB-D dataset of non-rigidly deforming scenes (\SCANS{} scenes, \FRAMES{} frames, \PAIRS{} densely aligned frame pairs).
    With this data, we propose a new method for non-rigid matching, which we integrate into a non-rigid reconstruction approach.
		}
		\label{fig:teaser}
	\end{center}    
}]

\input{0abstract.tex}

\input{1intro.tex}
\input{2prevwork.tex}

\input{3overview.tex}
\input{4method.tex}

\input{5dataset.tex}

\input{6results.tex}

\input{7conclusion.tex}

\section*{Acknowledgements}

\begin{footnotesize}

We would like to thank the expert annotators 
Sathya Ashok,
Omar Hedayat,
Haya Irfan,
Azza Jenane,
Soh Yee Lee, 
Suzana Spasova, and
Weile Weng
for their efforts in building the dataset. 
We further thank Edgar Tretschk for his help with data collection, Xiaochen Fan for his valuable contributions during his research visit, and Armen Avetisyan for numerous helpful discussions.
This work was supported by the Max Planck Center for Visual Computing and Communications (MPC-VCC), a TUM-IAS Rudolf M\"o{\ss}bauer Fellowship, the ERC Starting Grant \textit{Scan2CAD} (804724), the ERC Consolidator Grant \textit{4DRepLy} (770784), and the German Research Foundation (DFG) Grant \textit{Making Machine Learning on Static and Dynamic 3D Data Practical}.

\end{footnotesize}

\balance
{\small
\bibliographystyle{ieee_fullname}
\bibliography{egbib}
}

\newpage

\begin{appendix}
\input{appendix}
\end{appendix}

\end{document}

%% file: macros.tex
\def \OURS {{DeepDeform}}
\newcommand{\SCANS}{{400}}       %
\newcommand{\FRAMES}{{390,000}}  %
\newcommand{\MATCHES}{{149,228}}   %
\newcommand{\OCCLUSIONS}{{63,512}}   %
\newcommand{\MASKS}{{4,479}}   %
\newcommand{\PAIRS}{{5,533}}   %

\newcommand{\ru}    {\rule{0mm}{4mm}}

\definecolor{mncolor}{RGB}{255,50,00}
\newcommand\MATTHIAS[1] {\textbf{\textcolor{mncolor}{MN: #1}}}
\definecolor{abcolor}{RGB}{0,0,255}
\newcommand\AB[1] {\emph{\textcolor{abcolor}{AB: #1}}}
\definecolor{mzcolor}{RGB}{10,120,10}
\newcommand\MZ[1] {\textbf{\textcolor{mzcolor}{MZ: #1}}}
\definecolor{ctcolor}{RGB}{250,100,100}
\newcommand\CT[1] {\emph{\textcolor{ctcolor}{CT: #1}}}
\definecolor{mscolor}{RGB}{128,128,0}

%% file: 0abstract.tex
\begin{abstract}
Applying data-driven approaches to non-rigid 3D reconstruction has been difficult, which we believe can be attributed to the lack of a large-scale training corpus.
Unfortunately, this method fails for important cases such as highly non-rigid deformations.
We first address this problem of lack of data by introducing a novel semi-supervised strategy to obtain dense inter-frame correspondences from a sparse set of annotations.
This way, we obtain a large dataset of \SCANS{} scenes, over \FRAMES{} RGB-D frames, and \PAIRS{} densely aligned frame pairs; in addition, we provide a test set along with several metrics for evaluation.
Based on this corpus, we introduce a data-driven non-rigid feature matching approach, which we integrate into an optimization-based reconstruction pipeline.
Here, we propose a new neural network that operates on RGB-D frames, while maintaining robustness under large non-rigid deformations and producing accurate predictions.
Our approach significantly outperforms existing non-rigid reconstruction methods that do not use learned data terms, as well as learning-based approaches that only use self-supervision.
\blfootnote{\scriptsize Data / Benchmark: \url{https://github.com/AljazBozic/DeepDeform}}
\vfill
\end{abstract}

%% file: 1intro.tex
\vspace{-0.5cm}
\section{Introduction}
Non-rigid 3D reconstruction, i.e., the dense, space-time coherent capture of non-rigidly deforming surfaces in full temporal correspondence, is key towards obtaining 3D abstractions of the moving real world.
The wide availability of commodity RGB-D sensors, such as the Microsoft Kinect or Intel Realsense, has led to tremendous progress on static scene reconstruction methods.
However, robust and high-quality reconstruction of non-rigidly moving scenes with one depth camera is still challenging.
Applications for real-time non-rigid reconstruction range from augmented (AR) and virtual reality (VR) up to building realistic 3D holograms for fully immersive teleconferencing systems.
The seminal DynamicFusion \cite{newcombe2015dynamicfusion} approach was the first to show dynamic non-rigid reconstruction in real-time.
Extensions primarily differ in the used energy formulation.
Some methods use hand-crafted data terms based on dense geometry \cite{newcombe2015dynamicfusion,slavcheva2017killingfusion,slavcheva2018sobolevfusion}, dense color and geometry \cite{guo2017real,wang2018dynamic}, and sparse feature constraints \cite{innmann2016volumedeform}.
Other approaches leverage multi-camera RGB-D setups \cite{dou2016fusion4d,dou2017motion2fusion} for higher robustness.
However, there are very few reconstruction methods that use learning-based data terms for general real-world scenes rather than specific scenarios \cite{wei2016dense}, and that are trained to be robust under real-world appearance variation and difficult motions.
One reason for this is the lack of a large-scale training corpus.
One recent approach \cite{SchmidtNF17} proposes self-supervision for ground truth generation, i.e., they employ DynamicFusion \cite{newcombe2015dynamicfusion} for reconstruction and train a non-rigid correspondence descriptor on the computed inter-frame correspondences.
However, we show that existing non-rigid reconstruction methods are not robust enough to handle realistic non-rigid sequences; therefore this tracking-based approach does not scale to training data generation for real world scenes. 
Unfortunately, this means that self-supervision exactly fails for many challenging scenarios such as highly non-rigid deformations and fast scene motion.
By design, this approach trained with self-supervision cannot be better than the employed tracker.
We propose to employ semi-supervised training data by combining self-supervision with sparse user annotations to obtain dense inter-frame correspondences.
The annotated sparse point correspondences guide non-rigid reconstruction; this allows us to handle even challenging motions.
The result is a large dataset of \SCANS{} scenes, over \FRAMES{} RGB-D frames, and \PAIRS{} densely aligned frame pairs.
Based on this novel 
training corpus, we develop a new non-rigid correspondence matching approach (see Sec.~\ref{sec:NR_match}) that 
finds accurate matches between RGB-D frames and is robust to difficult real world deformations.
We further propose a re-weighting scheme that gives more weight to corner cases and challenging deformations during training.
Given a keypoint in a source frame, our approach predicts a probability heatmap of the corresponding location in the target frame. 
Finally, we integrate our learned data term into a non-rigid reconstruction pipeline that combines learned heatmap matches with a dense RGB-D reconstruction objective.
In addition, we introduce a new benchmark and metric for evaluating RGB-D based non-rigid 3D correspondence matching and reconstruction.
We extensively compare our new data-driven approach to existing hand-crafted features.
We also integrate the learned features into a non-rigid reconstruction framework, leading to significant improvement over state of the art.
\begin{figure}
    \vspace{-0.2cm}
    \includegraphics[width=\columnwidth]{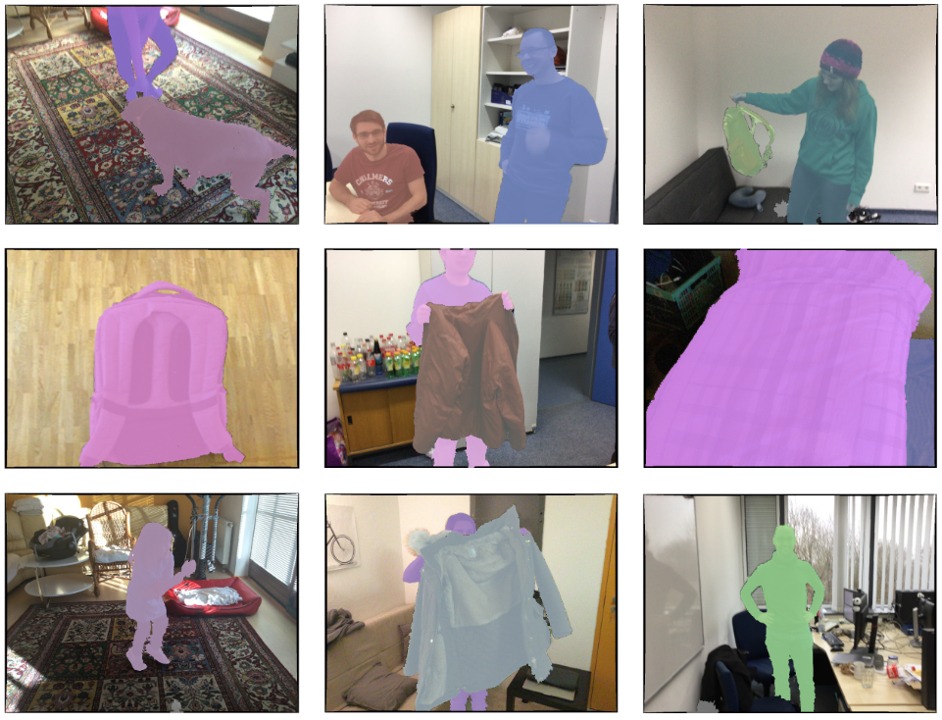}
    \vspace{-0.7cm}
    \caption{Our large-scale dataset contains a large variety of dynamic sequences with segmentation masks and point correspondences between different RGB-D frames.}
    \label{fig:dataset_overview}
\end{figure}
In sum, our contributions are:
\begin{itemize}
\item 
A semi-supervised labeling approach for dense non-rigid correspondence learning, resulting in a dataset featuring \SCANS{} annotated dynamic RGB-D sequences and \PAIRS{} densely aligned frame pairs.
\item 
A novel data-driven non-rigid correspondence matching strategy that leads to more robust correspondence estimation compared to the state-of-the-art hand-crafted and learned descriptors, especially in the case of extreme deformations. 
\item 
A non-rigid reconstruction approach for general scenes that combines learned and geometric data-terms 
and handles significantly faster and more complex motions than the state-of-the-art.
\end{itemize}

%% file: 2prevwork.tex
\section{Related Work}
Our approach is related to several research areas, such as volumetric 3D scene reconstruction, non-rigid object tracking, and learned correspondence matching.
We focus our discussion on the most related RGB-D based techniques.
For a detailed discussion, we refer to the recent survey~\cite{Zollhoefer2018RecoSTAR}.

\paragraph{Volumetric Scene Reconstruction}
Reconstructing static environments with a single RGB-D sensor has had a long history in vision and graphics, including KinectFusion \cite{newcombe:2011:KINFU,izadi11kinectfusion}, which employs a uniform voxel grid to represent the scene as a truncated signed distance function (TSDF) \cite{curless:1996:volumetric}, as well as many extensions to large-scale scenes~\cite{whelan2012kintinuous,chen:2013:SRV,steinbrucker2013large,niessner:2013:KINFUH}.
These techniques track the 6-DoF camera motion by solving a geometric model-to-frame alignment problem using a fast data-parallel variant of the point-to-plane Iterative Closest Point (ICP) algorithm \cite{rusinkiewicz:2001:EICP}.
Globally consistent reconstruction based on Bundle Adjustment \cite{zhou2013dense,choi2015robust} was for a long time only possible offline; data-parallel solvers now enable real-time frame rates \cite{dai2016bundle}. 
An alternative to TSDFs are point-based scene representations \cite{Keller2013,lefloch15anisotropic,lefloch:2017:curvature}.
Recent techniques also employ non-rigid registration to robustly handle loop closures \cite{whelan:2015:ElasticFusion,zhou13elastic}.

\paragraph{Non-Rigid Reconstruction}
The reconstruction of general non-rigidly deforming objects based on real-time scan data has a long tradition \cite{wand2009efficient}.
One class of methods uses pre-defined templates, e.g., human templates, to capture pose and time-varying shape of clothed humans from RGB-D~\cite{DoubleFusion} or stereo camera data~\cite{DBLP:journals/tog/WuSVT13}.
First template-less approaches had slow offline runtimes and only worked for slow and simple motions.
The approach of \cite{dou20153d} solves a global optimization problem to reconstruct the canonical shape of a non-rigidly deforming object given an RGB-D video sequence as input, but does not recover the time-dependent non-rigid motion across the entire sequence.
The first approach to demonstrate truly dynamic reconstruction of non-rigid deformation and rest shape in real-time was DynamicFusion~\cite{newcombe2015dynamicfusion}.
Since this seminal work, many extensions have been proposed.
VolumeDeform~\cite{innmann2016volumedeform} improves tracking quality based on sparse feature alignment.
In addition, they parameterize the deformation field based on a dense volumetric grid instead of a sparse deformation graph. %
The KillingFusion \cite{slavcheva2017killingfusion} and SobolevFusion \cite{slavcheva2018sobolevfusion} approaches allow for topology changes, but do not recover dense space-time correspondence along the complete input sequence.
Other approaches jointly optimize for geometry, albedo, and motion \cite{guo2017real} to obtain higher robustness and better quality.
The approach of Wang et al.~\cite{wang2018dynamic} employs global optimization to minimize surface tracking errors.
In contrast to these methods using a single RGB-D camera, other techniques use multiple color~\cite{deAguiar:2008,Vlasic:2008} or depth cameras~\cite{Ye2012KinectsMocap,DBLP:journals/corr/WangWVHCML16,dou2016fusion4d,dou2017motion2fusion}, which enables high-quality reconstruction at the cost of more complex hardware.
We propose a new learning-based correspondence matching and reconstruction approach that outperforms existing techniques.

\paragraph{Learning Rigid Correspondence Matching}
Historically, correspondence matching for the task of rigid registration has been based on hand-crafted geometry descriptors~\cite{Johnson:1999,Frome2004,Tombari:2010,Rusu2008,Rusu:2009}.
If color information is available in addition to depth, SIFT \cite{Lowe:2004} or SURF \cite{Bay:2008} can be used to establish a sparse set of feature matches between RGB-D frames.
More recently, 2D descriptors for feature matching in static scenes have been learned directly from large-scale training corpora \cite{Simo-SerraTFKFM15,Simonyan2014,Leung2015,YiTLF16,Jure2015}. %
The Matchnet \cite{Leung2015} approach employs end-to-end training of a CNN to extract and match patch-based features in 2D image data.
Descriptors for the rigid registration of static scenes can be learned and matched directly in 3D space with the 3DMatch~\cite{zeng20163dmatch} architecture.
Visual descriptors for dense correspondence estimation can be learned in a self-supervised manner by employing a dense reconstruction approach to automatically label correspondences in RGB-D recordings \cite{SchmidtNF17}.
Descriptor learning and matching for static scenes has been well-studied, but is lacking in the challenging non-rigid scenario.
While class-specific dense matching of non-rigid scenes has been learned for specific object classes \cite{wei2016dense,groueix2018b}, none of these techniques can handle arbitrary deforming non-rigid objects.
We believe one reason for this is the lack of a large-scale training corpus.
In this work, we propose such a corpus and demonstrate how non-rigid matching between depth images can be learned end-to-end.

\paragraph{RGB-D Datasets}
While we have seen a plethora of RGB-D datasets for static scenes, such as NYU \cite{Silberman:ECCV12}, SUN RGB-D \cite{Song_2015_CVPR}, and ScanNet \cite{dai2017scannet}, the largest of which have thousands of scans, non-rigid RGB-D datasets remain in their infancy.
While these datasets can be used to pretrain networks for the task of non-rigid correspondence matching, they do not capture the invariants that are useful for the much harder non-rigid setting, and thus lead to sub-par accuracy.
Current non-rigid reconstruction datasets are far too small and often limited to specific scene types~\cite{deAguiar:2008,Ye2012KinectsMocap}, which is not sufficient to provide the required training data for supervised learning.
The datasets that provide real-world depth recordings \cite{guo2017robust,innmann2016volumedeform} do not come with ground truth reconstructions, which makes objectively benchmarking different approaches challenging.
Other datasets that are commonly used for evaluation do not provide real-world depth data, e.g.,~\cite{deAguiar:2008,Vlasic:2008}.
In this work, we introduce the first large-scale dataset for non-rigid matching based on semi-supervised labeling and provide a benchmark enabling objective comparison of different approaches.

%% file: 4method.tex
\section{Data-driven Non-Rigid Matching}
\label{sec:NR_match}
Our goal is to find matches between a source and a target RGB-D frame.
To this end, we propose a network architecture for RGB-D matching based on a Siamese network \cite{KumarCR15} with two towers.
Input to the network are two local patches of size $224 \times 224$ pixels each (with 3 color channels and 3-dimensional points in camera coordinate space).
We assume that the source patch is centered at a feature point.
\begin{figure*}
\vspace{-0.5cm}
\includegraphics[width=\linewidth]{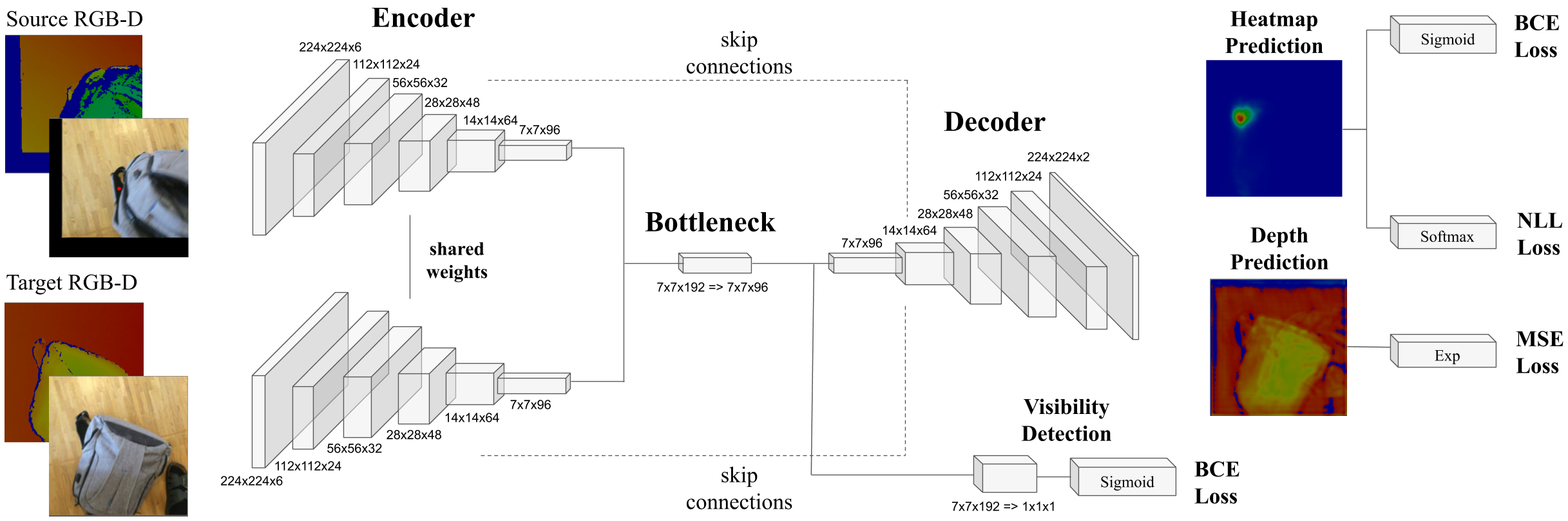} \\
\vspace{-0.5cm}
\caption{
We devise an end-to-end architecture for RGB-D matching based on a Siamese network to find matches between a source and a target frame.
Our network is based on two towers that share the encoder and have a decoder that predicts two probability heatmaps in the target frame that encode the likelihood of the location of the source point.
Our network also predicts a depth value for the matched point and a visibility score that measures if the source point is visible in target frame.
}
\label{fig:network-architecture}
\end{figure*}

\paragraph{Heatmap}
The goal is to predict a probability heatmap $\mathcal{H}$ in the target frame that gives the likelihood of the location of the source point.
First, we compute a \emph{sigmoid-heatmap}:
$$\mathcal{H}_{\textrm{sg}} = \sigma_{\textrm{sg}}\big(\mathbf{H}(D_{\textrm{out}})\big) \enspace{.}$$
It is computed based on a sigmoid activation $\sigma_{\textrm{sg}}$ to map responses to $[0,1]$.
Here, $D_{\textrm{out}}$ is the output feature map of the last layer of the decoder and $\mathbf{H}$ is a convolutional layer converting feature space into heatmap values.
This is equivalent to independent binary classification problems per pixel.
Second, we also compute a \emph{softmax-heatmap}:
$$\mathcal{H}_{\textrm{sm}} = \sigma_{\textrm{sm}}\big(\mathbf{H}(D_{\textrm{out}})\big) \enspace{.}$$
Here we use a softmax activation $\sigma_{\textrm{sm}}$ to make the complete heatmap a probability distribution, i.e., it sums to one.
As ground truth for the heatmap prediction we could take an image with zero values everywhere except at the ground truth pixel position that we set to one.
To prevent the trained network from predicting only zero values,
we apply a Gaussian kernel $G_{x_{gt}}$ around the ground truth pixel. It sets the ground truth pixel's value to one and decays the neighboring pixel values to zero with standard deviation of $7$ pixels, resulting in the ground truth heatmap $\mathcal{H}_{\textrm{gt}}$. 
We also add larger weight to pixels close to the ground truth pixel, defining the pixel weight as $w_\mathcal{H}(x) = 1 + 10 \cdot G_{x_{gt}}(x)$.
The heatmap loss is then computed as:
\begin{align*}
    \mathcal{L}_\mathcal{H} = &\sum_i{\Phi_{bce}( w_\mathcal{H} (\mathcal{H}_{\textrm{sg}} - \mathcal{H}_{\textrm{gt}} ))} + \\
    &\lambda_\textrm{nll} \sum_i{\Phi_{nll}( w_\mathcal{H} (\mathcal{H}_{\textrm{sm}} - \mathcal{H}_{\textrm{gt}} ))}  \enspace{.}
\end{align*}
Here, $\Phi_{bce}(\bullet)$ denotes the binary cross entropy loss and $\Phi_{nll}(\bullet)$ the negative log-likelihood loss, and we empirically determined a weight $\lambda_\textrm{nll} = 10$.
From these two probability heatmaps, a single one is computed as $\mathcal{H} = \mathcal{H}_{sg} \otimes \mathcal{H}_{sm}$, where $\otimes$ is the Hadamard product.

\paragraph{Depth}
In addition to heatmap prediction, our network also predicts the matched point's depth value in the target camera's coordinate system. Inspired by \cite{mehta2017vnect}, we predict the depth densely, predicting the same depth value for every pixel in the output image:
$$\mathcal{D} = \exp\big(\mathbf{D}(D_{\textrm{out}})\big) \enspace{.}$$
Here, $\mathbf{D}$ is a convolutional layer converting feature space into depth values, and the exponential is applied to guarantee positive depth predictions.
Ground truth for depth prediction $\mathcal{D}_{\textrm{gt}}$ is the depth of the ground truth match, repeated for the whole image. 
Since we want to encourage depth prediction to focus on the matched pixel, we again use pixel weighting, this time in the form of $w_\mathcal{D}(x) = G_{x_{gt}}(x)$, setting the center pixel's weight to $1$ and decaying the weights to $0$.
Using the weighted version of mean squared error $\Phi_{mse}(\bullet)$ we employ the following loss for depth prediction:
\begin{equation*}
    \mathcal{L}_\mathcal{D} = \lambda_\textrm{d} \sum_i{\Phi_{mse}( w_\mathcal{D} (\mathcal{D} - \mathcal{D}_{\textrm{gt}}) )}  \enspace{.}
\end{equation*}

\paragraph{Visibility}
Furthermore, we also predict a visibility score $\in [0, 1]$ that measures whether the source point is visible (high value) or occluded (low value) in the target frame:
$$\mathcal{V} = \sigma_{\textrm{sg}}\big(\mathbf{V}(B_{\textrm{out}})\big) \enspace{.}$$
Here, $B_{\textrm{out}}$ is the output feature map of the bottleneck layer, $\mathbf{V}$ is a convolutional layer, and $\sigma_{\textrm{sg}}$ a sigmoid activation.
The visibility loss takes the following form:
\begin{equation*}
    \mathcal{L}_\mathcal{V} = \sum_i{\Phi_{bce}( \mathcal{V} - \mathcal{V}_{\textrm{gt}} )} \enspace{.}
\end{equation*}

In the end, we train the network using a weighted combination of all presented loss functions:
\begin{equation*}
    \mathcal{L} = \mathcal{L}_\mathcal{H} + \lambda_\mathcal{D} \mathcal{L}_\mathcal{D} + \lambda_\mathcal{V} \mathcal{L}_\mathcal{V}  \enspace{.}
\end{equation*}
In all experiments we use the constant and empirically determined weights $\lambda_\mathcal{D} = 100$ and $\lambda_\mathcal{V} = 1$. An overview of the network architecture is given in Fig.~\ref{fig:network-architecture}. More network and training details can be found in the supplemental.

\section{Non-Rigid Reconstruction Pipeline} \label{sec:reconstructionp}
We integrate the learned non-rigid matching algorithm into a
non-rigid RGB-D reconstruction framework that efficiently tracks dense, space-time coherent, non-rigid deformations on the GPU and also provides an efficient volumetric fusion backend.
A canonical model of the scene is reconstructed from data, in parallel to tracking the non-rigid deformations, and stored based on a truncated signed distance field (TSDF) represented by a uniform voxel grid.
New observations are fused into the grid based on an exponentially moving average.
The non-rigid scene motion is tracked based on the following tracking energy:
\begin{equation*}
E_\textrm{total}(\mathcal{T}) = E_\textrm{data}(\mathcal{T}) + \lambda_\textrm{learned} E_\textrm{learned}(\mathcal{T}) + \lambda_\textrm{reg} E_\textrm{reg}(\mathcal{T}) \enspace{.}
\end{equation*}
The weights $\lambda_\textrm{learned}=1$ and $\lambda_\textrm{reg}=1$ are empirically determined and balance the different terms.

\subsection{Deformation Model}
To parameterize scene motion, similar to \cite{Sumner:2007}, we employ a coarse deformation graph $\mathcal{G}$ with $K$ deformation nodes $\mathbf{g}_i \in \mathbb{R}^3$.
The graph models the deformation of space based on the convex combination of local per-node transformations that are parameterized by rotation parameters $\mathbf{\theta}_i \in \mathbb{R}^{3}$ in Lie algebra space and a translation vector $\mathbf{t}_i \in \mathbb{R}^3$.
In total, this leads to $2k$ free variables to describe scene motion that we jointly refer to as $\mathcal{T}$.
This enables us to decouple the number of free variables from the complexity of the reconstructed scene.
The deformation nodes are connected based on proximity, for details we refer to the original embedded deformation paper \cite{Sumner:2007}.

\subsection{Optimization Terms}
For data term $E_\textrm{data}(\mathcal{T})$, similar to \cite{newcombe2015dynamicfusion,innmann2016volumedeform}, we employ dense point-to-point and point-to-plane alignment constraints between the input depth map and the current reconstruction.
For regularizer $E_\textrm{reg}$, we employ the as-rigid-as-possible (ARAP) constraint~\cite{Sorkine:2007} to enforce locally rigid motion.
In addition, we integrate a sparse feature alignment term based on our learned correspondences (see Sec.~\ref{sec:NR_match}).
For each node $\mathbf{g}_i$ of the current deformation graph, we predict a probability heatmap $\mathcal{H}_i$ that gives the likelihood of its 2D uv-position in the current input depth map, using the initial depth map as the reference frame.
Furthermore, we also back-project the pixel with the maximum heatmap response into a 3D point $\mathbf{p}_i \in \mathbb{R}^3$, using its depth.
We aim to align the graph nodes with the maximum in the corresponding heatmap using the following alignment constraint:
\begin{align*}
    E_{\textrm{learned}}(\mathcal{T}) = &\sum_{g_i \in \mathcal{G}}{ \big( 1 - \mathcal{H}_i( \pi( \mathbf{g}_i + \mathbf{t}_i ) ) \big)^2 } + \\  &\lambda_{point} \sum_{g_i \in \mathcal{G}}{ \big( \mathbf{g}_i + \mathbf{t}_i - \mathbf{p}_i \big)^2 } \enspace{.}
\end{align*}
Here, $\pi : \mathbb{R}^3 \rightarrow \mathbb{R}^2$ is the projection from 3D camera space to 2D screen space. The heatmap $\mathcal{H}_i$ is normalized to a maximum  of $1$. 
We empirically set $\lambda_{point} = 10$.
In order to handle outliers, especially in the case of occluded correspondences, we make use of the predicted visibility score and the predicted depth value of the match. We filter out all heatmap correspondences with visibility score  $< 0.5$. We compare the predicted depth with the queried depth from the target frame's depth map at the pixel with the maximum heatmap response and invalidate any correspondences with a depth difference  $> 0.15$ meters.

\subsection{Energy Optimization}
We efficiently tackle the underlying optimization problem using a data-parallel Gauss-Newton solver to find the deformation graph $\mathcal{G}^*$ that best explains the data:
\begin{equation*}
   \mathcal{G}^* = \argmin{E_\textrm{total}(\mathcal{G})} \enspace{.}
\end{equation*}
In the Gauss-Newton solver, we solve the underlying sequence of linear problems using data-parallel preconditioned conjugate gradient (PCG).
For implementation details, we refer to the supplemental document.

%% file: 5dataset.tex
\section{Semi-supervised Data Acquisition} \label{sec:dataset}
In the following, we provide the details of our semi-supervised non-rigid data collection process that is used for training the non-rigid matching and for the evaluation of non-rigid reconstruction algorithms.
The high-level overview of the data acquisition pipeline is shown in Fig.~\ref{fig:teaser}.

\subsection{Data Acquisition}
In order to obtain RGB-D scans of non-rigidly moving objects, we use a Structure Sensor mounted on an iPad.
The depth stream is recorded at a resolution of $640 \times 480$ and $30$ frames per second; the RGB stream is captured with the iPad camera at a resolution of $1296 \times 968$ pixels that is calibrated with respect to the range sensor.
Regarding the scanning instructions, we follow the ScanNet~\cite{dai2017scannet} pipeline.
However, in our case, we focus on scenes with one up to several non-rigidly moving objects in addition to a static background.
In total, we recorded \SCANS{} scenes with over \FRAMES{} RGB-D frames.

\subsection{Data Annotation}
We crowd sourced sparse ground truth correspondence annotations and segmentation masks for our novel data set.
To this end, we employed a web-based annotation tool.
The annotation was divided into two tasks.
Firstly, we select up to 10 frames per sequence.
All dynamic objects that are found in these frames are given unique instance ids (the same instance in the whole sequence) and their masks are annotated in each frame.
To accelerate mask segmentation, we use a hierarchy of superpixels as candidate brush sizes.
Secondly, among the annotated frames up to 10 frame pairs are selected, and the sparse correspondences between all dynamic objects are annotated.
Expert annotators were instructed to annotate correspondences uniformly over the complete object, labeling about $20$ point matches per frame pair.
Furthermore, in parts of the source image that are occluded in the target image occlusion points were uniformly selected to collect data samples for visibility detection.
The dynamic object segmentation task takes on average about $1$ min per frame, while the correspondence labeling task takes on average about $2$ min per frame.

\subsection{Dense Data Alignment}
Using the annotated object masks and the sparse matches, dense non-rigid alignment of the frame pairs is performed. 
We follow a similar approach as for non-rigid reconstruction (see Sec.~\ref{sec:reconstructionp}), based on the sparse deformation graph of \cite{Sumner:2007}.
The deformation graph is defined on the source frame's depth map, covering only the dynamic object by using the source object mask.
The final energy function that is optimized to align the source RGB-D frame to the target RGB-D frame is:
\begin{align*}
E_\textrm{total}(\mathcal{T}) = &E_\textrm{data}(\mathcal{T}) + \lambda_\textrm{photo} E_\textrm{photo}(\mathcal{T}) + \lambda_\textrm{silh} E_\textrm{silh}(\mathcal{T})+ \\
&\lambda_\textrm{sparse} E_\textrm{sparse}(\mathcal{T}) + \lambda_\textrm{reg} E_\textrm{reg}(\mathcal{T}) \enspace{.}
\end{align*}
Here, $E_\textrm{photo}(\mathcal{T})$ encourages the color gradient values from the source frame to match the target frame, $E_\textrm{silh}$ penalizes deformation of the object outside of the target frame's object mask, and $E_\textrm{sparse}$ enforces annotated sparse matches to be satisfied.
The weights $\lambda_\textrm{photo}=0.001$, $\lambda_\textrm{silh}=0.0001$, $\lambda_\textrm{sparse}=100.0$ and $\lambda_\textrm{reg}=10.0$ have been empirically determined.
Details about the different optimization terms and a qualitative comparison of their effects can be found in the supplemental document.
In order to cope with simple apparent topology changes that are very common while capturing natural non-rigid motion, such as the hand touching the body in one frame and moving away in another frame, we execute non-rigid alignment in both directions and compute the final non-rigid alignment using forward-backward motion interpolation, similar to \cite{zampogiannis2019topology}. 
At the end, a quick manual review step is performed, in which, if necessary, any incorrectly aligned mesh parts are removed.
The review step takes about $30$ seconds per frame.
Examples of dense alignment results and the employed review interface can be found in the accompanying video.

%% file: 6results.tex
\section{Experiments} \label{sec:results}
We provide a train-val-test split with 340 sequences in the training set, 30 in the test set, and 30 in the validation set.
We made sure that there is no overlap between captured environments between training and validation/test scenes.

\subsection{Non-Rigid Matching Evaluation}
For a given set of pixels (and corresponding 3D points) in the source image, the task is to find the corresponding pixel (and 3D point) in the target image.
We evaluate the average 2D pixel and 3D point error (in meters), and compute the matching accuracy (ratio of matches closer than $20$ pixels or $0.05$ meters from the ground truth correspondences).
We compare our non-rigid matching approach to several hand-crafted feature matching strategies, that are based on depth or color based descriptors, and  to the learned 3Dmatch~\cite{zeng20163dmatch} descriptor, see Tab.~\ref{tab:matching-comparison}. 
Specifically, we compare to the hand-crafted geometry descriptors, such as Unique Signatures of Histograms (SHOT) \cite{Tombari:2010} and the Fast Point Feature Histograms (FPFH) \cite{Rusu:2009}.
We also compare to color-based descriptors, e.g., SIFT \cite{Lowe:2004} and SURF \cite{Bay:2008}, that can be used to establish a sparse set of matches been RGB-D frames.
Finally, we train a learned descriptor from \cite{zeng20163dmatch}, patch-based random forest matcher from \cite{wang2016global} and optical flow prediction network from \cite{ilg2017flownet2} on our training sequences.
Our method consistently outperforms all the baselines.

\begin{table}[]
\footnotesize
\centering
\begin{tabular}{|l||c|c||c|c|}
 \hline
\textbf{Method} & \textbf{2D-err} & \textbf{3D-err} & \textbf{2D-acc} & \textbf{3D-acc}  \\ \hline
SIFT~\cite{Lowe:2004} & $138.40$ & $0.552$ & $16.20$ & $14.08$ \\ 
SURF~\cite{Bay:2008} & $125.72$ & $0.476$ & $22.13$ & $19.82$ \\ 
SHOT~\cite{Tombari:2010} & $105.34$ & $0.342$ & $13.43$ & $11.51$ \\
FPFH~\cite{Rusu:2009} & $109.49$ & $0.393$ & $10.85$ & $9.43$ \\
3Dmatch~\cite{zeng20163dmatch} & $68.98$ & $0.273$ & $30.50$ & $25.33$ \\
GPC~\cite{wang2016global}& $65.04$ & $0.231$ & $31.93$ & $28.16$ \\
FlowNet-2.0~\cite{ilg2017flownet2}& $27.32$ & $0.118$ & $68.68$ & $63.67$ \\ \hline
Ours-$12.5\%$ & $78.82$ & $0.268$ & $27.17$ & $23.28$ \\
Ours-$25.0\%$ & $58.28$ & $0.197$ & $40.32$ & $35.81$ \\ 
Ours-$50.0\%$ & $45.43$ & $0.156$ & $50.70$ & $46.57$ \\ \hline
Ours-Rigid & $57.87$ & $0.270$ & $40.93$ & $35.64$ \\
Ours-SelfSupervised & $33.34$ & $0.121$ & $60.87$ & $55.70$ \\
Ours-Sparse & $31.42$ & $0.106$ & $58.53$ & $52.24$ \\ \hline
Ours-NoWeighting & $23.72$ & $0.083$ & $73.13$ & $68.46$ \\ \hline
Ours & $\mathbf{19.56}$ & $\mathbf{0.073}$ & $\mathbf{77.60}$ & $\mathbf{72.48}$ \\ \hline
\end{tabular}
\vspace{-0.2cm}
\caption{\label{tab:matching-comparison} We outperform all baseline matching methods by a considerable margin. 2D/3D errors are average pixel/point errors, and 2D/3D accuracy is the percentage of pixels/points with distance of at most $20$ pixels/$0.05$ meters.}
\end{table}

\subsection{Non-Rigid Reconstruction Results}
\begin{figure}
    \includegraphics[width=\columnwidth]{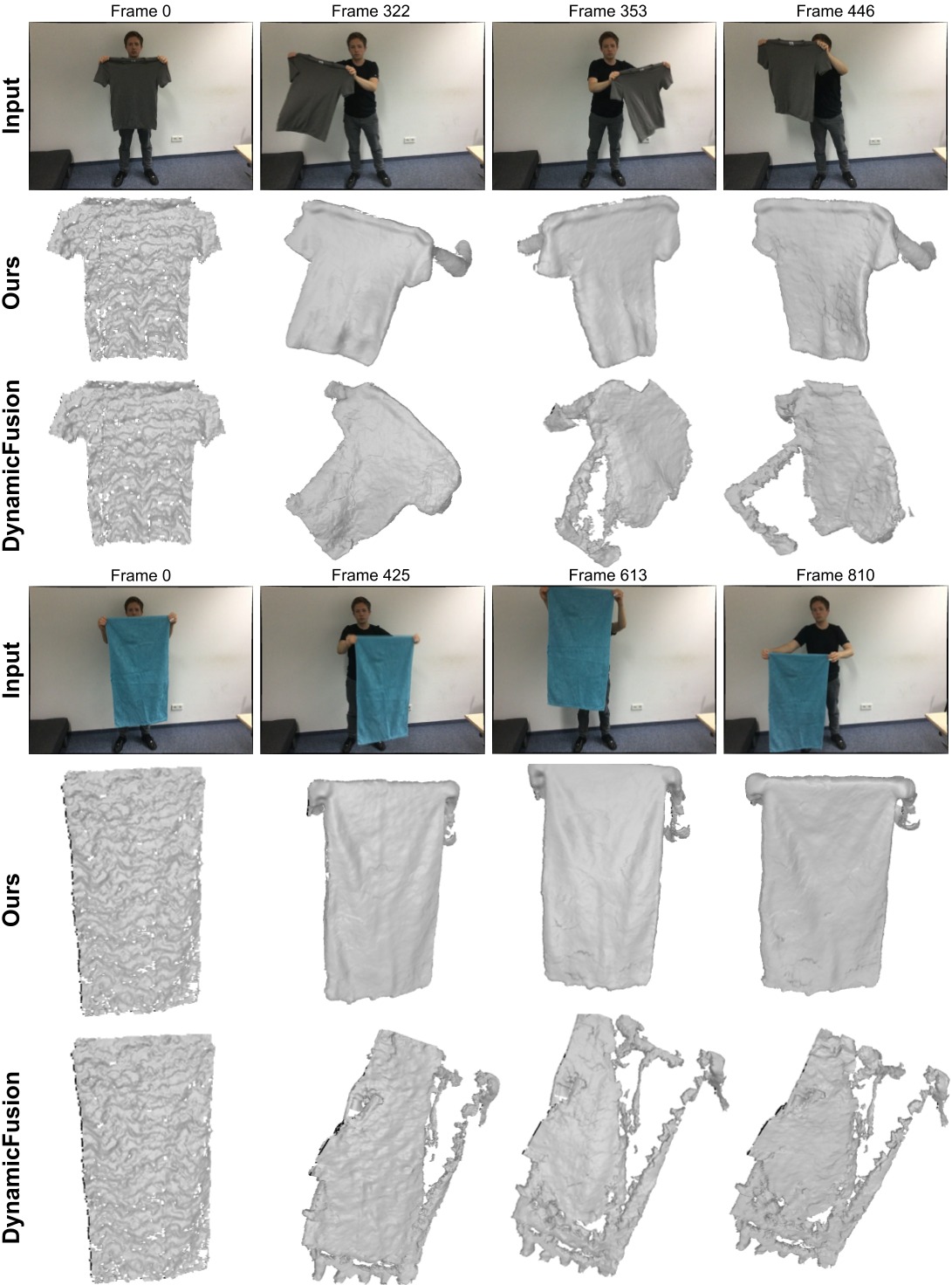}
    \vspace{-0.6cm}
    \caption{Qualitative comparison to DynamicFusion (reimplementation). \vspace{-0.3cm}
    }
    \label{fig:qualitative_comparison_dynamicfusion}
\end{figure}

\begin{figure}
    \includegraphics[width=\columnwidth]{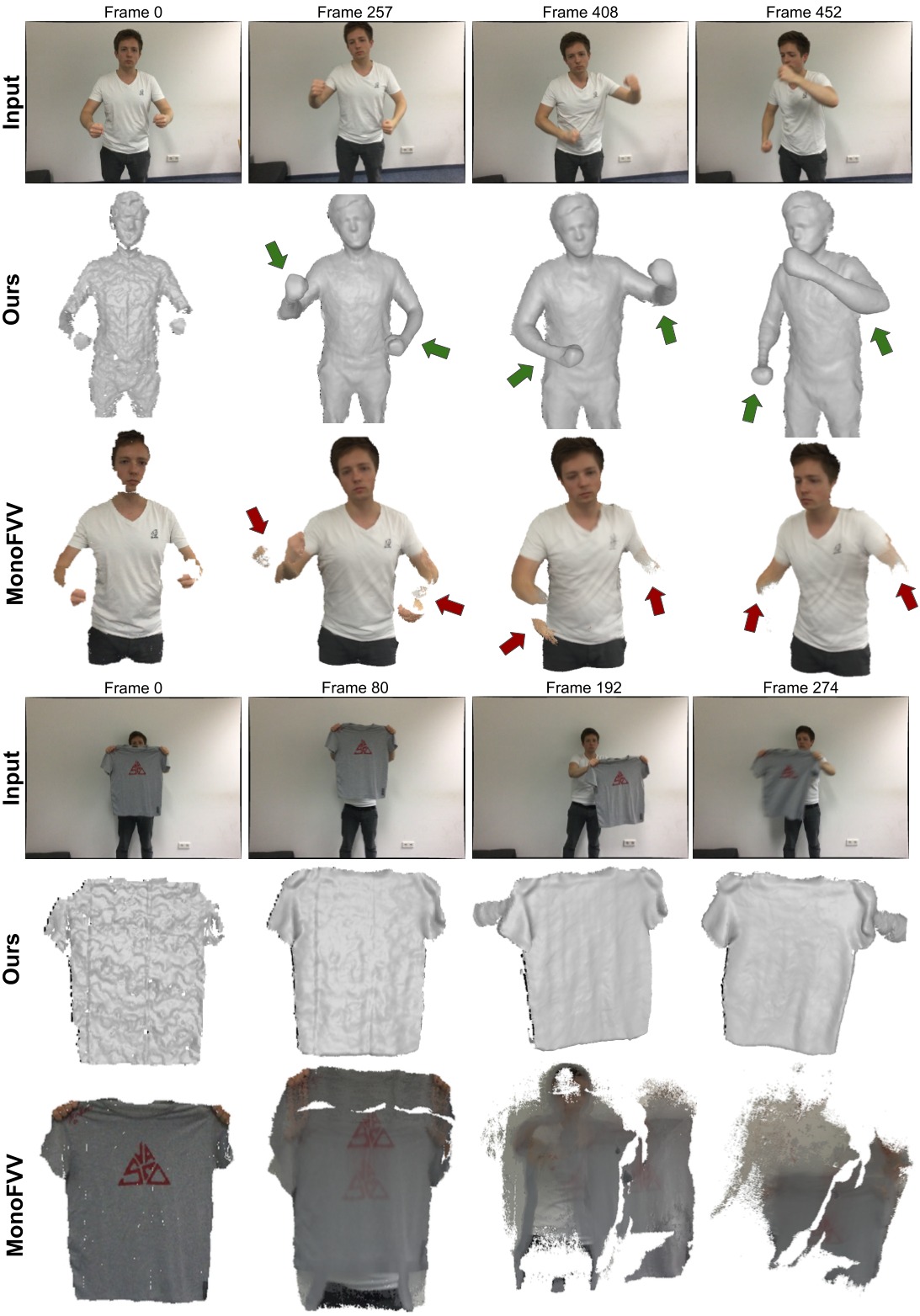}
    \vspace{-0.5cm}
    \caption{Qualitative comparison of reconstruction results of our method and MonoFVV~\cite{guo2017real}. Reconstruction results were kindly provided by the authors.
    }
    \label{fig:qualitative_comparison_monofvv}
\end{figure}

We integrated our learned matching strategy into a non-rigid reconstruction pipeline.
Our learned data term significantly improves reconstruction quality, both qualitatively and quantitatively. %
To be able to perform a quantitative comparison on our test sequences, we used our re-implementation of \cite{newcombe2015dynamicfusion}, and the code or results provided from the authors of \cite{innmann2016volumedeform, slavcheva2017killingfusion, slavcheva2018sobolevfusion, guo2017real}.
We also replaced our data-driven correspondence matching module with the descriptor learning network from \cite{zeng20163dmatch}, trained on our data, and used it in combination with 3D Harris keypoints.
The quantitative evaluation is shown in Tab.~\ref{tab:reconstruction-comparison}.
The evaluation metrics measure deformation error (a 3D distance between the annotated and computed correspondence positions) and geometry error (comparing depth values inside the object mask to the reconstructed geometry).
Deformation error is the more important metric, since it also measures tangential drift within the surface.
To be able to know which dynamic object to reconstruct if multiple are present, we always provide the initial ground truth segmentation mask of the selected object.
All approaches in Tab.~\ref{tab:reconstruction-comparison} were evaluated on all $30$ test sequences to provide a comparison on different kinds of objects and deformable motions.
\cite{guo2017real} provided results on two challenging test sequences, their average deformation and geometry error are $21.05$ cm and $14.87$ cm respectively, while our approach achieves average errors of $3.63$ cm and $0.48$ cm. 
Our approach outperforms the state of the art by a large margin.
The methods \cite{slavcheva2017killingfusion} and \cite{slavcheva2018sobolevfusion} do not compute explicit point correspondences from the canonical frame to other frames, so we could not evaluate these approaches quantitatively; we provide qualitative comparison on our sequences in the supplemental document.
We also show qualitative comparisons with our re-implementation of DynamicFusion~\cite{newcombe2015dynamicfusion} in Fig.~\ref{fig:qualitative_comparison_dynamicfusion} and with the state-of-the-art approach of \cite{guo2017real} in Fig.~\ref{fig:qualitative_comparison_monofvv}.
Our learned correspondences enable us to handle faster object motion as well as challenging planar motion, where even photometric cues fail, for instance due to uniform object color.

\subsection{Ablation Study}
\begin{figure}
    \centering
    \includegraphics[width=\columnwidth]{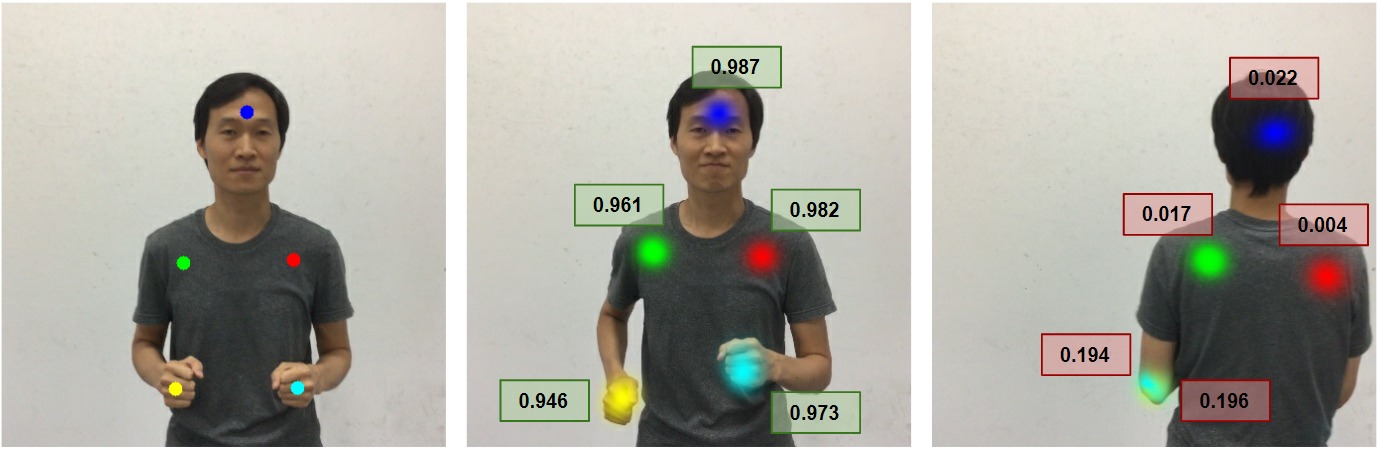}
    \vspace{-0.6cm}
    \caption{Visibility detection enables filtering of network correspondences that are occluded. The visibility score is in the range $[0, 1]$, it is high for visible correspondences and low for occluded parts. We filter out all correspondences with visibility score less than $0.50$. }
    \label{fig:occlusion-detection}
\end{figure}

We evaluated different components of our network and their effect on the reconstruction quality, see Tab.~\ref{tab:reconstruction-comparison}. 
Since some sequences include motions in which large parts of the reconstructed object are occluded, as can be observed in Fig.~\ref{fig:occlusion-detection}, using visibility detection for correspondence pruning makes our method more robust.
Furthermore, since depth measurements and heatmap predictions can both be sometimes noisy, adding correspondence filtering with depth prediction further improves the reconstruction results.
\begin{table}[]
\footnotesize
\centering
\begin{tabular}{|l|c|c|c|}
\hline
\textbf{Method} & \textbf{Def. error (cm)} & \textbf{Geo. error (cm)} \\ \hline
DynamicFusion re-impl. \cite{newcombe2015dynamicfusion} & $6.31$ & $1.08$ \\ 
VolumeDeform \cite{innmann2016volumedeform} & $21.27$ & $7.78$ \\
DynamicFusion + 3Dmatch & $6.64$ & $1.59$ \\ \hline
Ours-Rigid & $12.21$ & $2.30$ \\
Ours-Sparse & $8.24$ & $0.77$ \\
Ours-SelfSupervised & $5.47$ & $0.54$ \\ \hline
Ours-Base & $3.94$ & $0.43$ \\
Ours-Occlusion & $3.70$ & $0.42$ \\
Ours-Occlusion+Depth & $\mathbf{3.28}$ & $\mathbf{0.41}$ \\ \hline
\end{tabular}
\vspace{-0.3cm}
\caption{\label{tab:reconstruction-comparison} Comparison with state-of-the-art approaches. Our learned correspondences significantly improve both tracking and reconstruction quality. We also provide ablation studies on training data type and different network parts.}
\end{table}

\subsection{Data Generation Evaluation}
\begin{figure}
    \centering
    \includegraphics[width=\columnwidth]{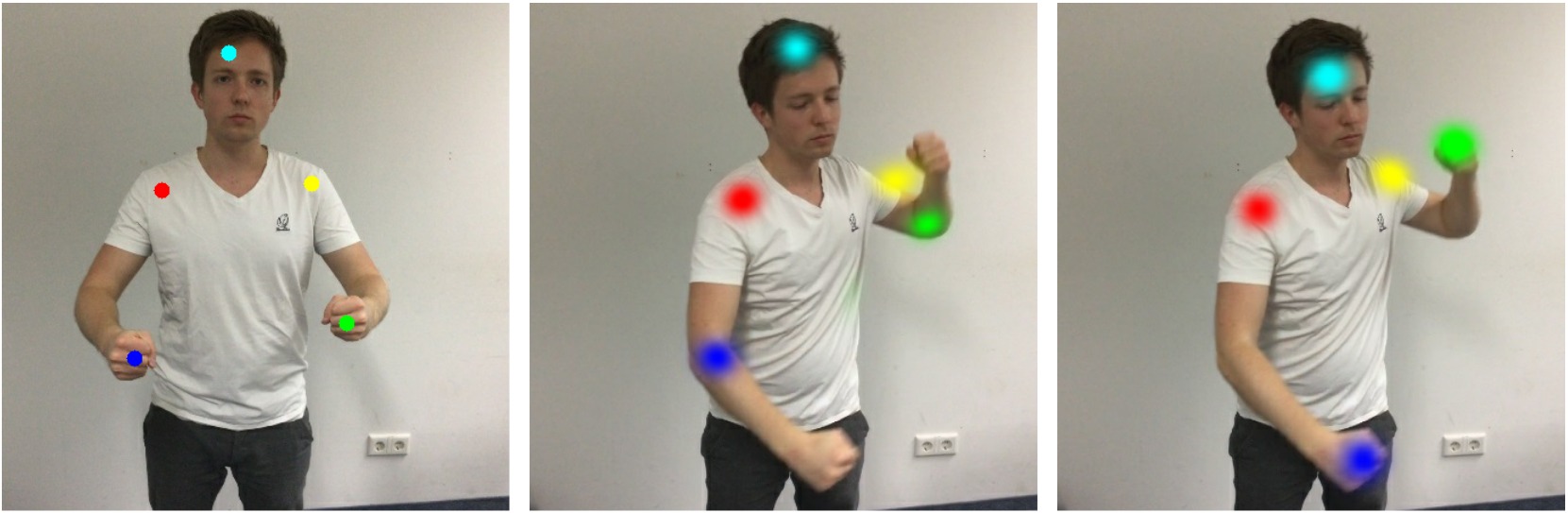}
    \vspace{-0.6cm}
    \caption{Comparison of correspondence prediction for reference frame (left) using self-supervised training data (middle) and semi-supervised dense data (right).}
    \label{fig:selfsupervised-vs-dense}
\end{figure}

To show the importance of our semi-supervised strategy for constructing the training corpus, we evaluate how different training corpora influence the performance of data-driven reconstruction methods.
Aside from our training data, which has been generated using dense semi-supervised frame alignment, we used a publicly available rigid dataset of indoor scenes (from \cite{dai2017scannet}), self-supervised alignment of our sequences (as in \cite{SchmidtNF17}), and only manually annotated sparse samples from our dataset.
We provide comparison on both non-rigid matching in Tab.~\ref{tab:matching-comparison} and non-rigid reconstruction in Tab.~\ref{tab:reconstruction-comparison}.
Using only rigid data does not generalize to non-rigid sequences.
While sparse matches already improve network performance, there is not enough data for reliable correspondence prediction on every part of the observed object.
In addition, annotated sparse matches are usually matches on image parts that are easy for humans to match, and there are much less accurate correspondences in areas with uniform color.
In the self-supervised setting, the network gets dense correspondence information, which improves the method's reconstruction performance compared to using only sparse features.
However, without semi-supervised densely aligned frame pairs, we can only generate matches for the simple deformations, where the DynamicFusion approach can successfully track the motion.
Therefore, the performance of the network trained only on self-supervised data degrades considerably on more extreme deformations, as can be seen in Fig.~\ref{fig:selfsupervised-vs-dense}.
Dense alignment of too far away frames is needed for accurate network prediction also in the case of extreme deformations.
Since the majority of the dense aligned matches is still moving rigidly, it turned out to be beneficial to sample more deformable samples during training.
In order to estimate which parts of the scene are more deformable, we employed sparsely annotated matches and ran RANSAC in combination with a Procrustes algorithm to estimate the average rigid pose.
The more the motion of each sampled match differs from the average rigid motion, the more often we sample it during network training using a multinomial distribution of the non-rigid displacement weights.
This strategy improved the network performance, as is shown in Tab.~\ref{tab:matching-comparison}, compared to training on non-weighted samples.
Finally, we demonstrate how much data is needed to achieve robust correspondence prediction performance; i.e., using less training data considerably degrades matching accuracy, as summarized in Tab.~\ref{tab:matching-comparison}, where we trained networks using only $12.5\%$, $25.0\%$, and $50.0\%$ of the training data.

%% file: 7conclusion.tex
\subsection{Limitations}
While learned correspondences make tracking of fast motion more robust, there is still room for improvement when reconstructing dynamic objects.
One pressing issue is that background clutter might be accidentally fused with the object when the object is close to the background.
In this case, the reconstructed shape would slowly grow and we might also start reconstructing the background.
This can cause wrong deformation graph connectivity and lead to tracking failures.
A potential future avenue is to subtract and ignore the background; e.g., we could use our annotated object masks to develop a data-driven method.

\section{Conclusion}
We have proposed a neural network architecture for matching correspondences in non-rigid sequences that operates on RGB-D frames and demonstrated that our learned descriptors outperform existing hand-crafted ones.
In addition, we introduced the first large-scale dataset that is composed of \SCANS{} scenes, over \FRAMES{} RGB-D frames, and \PAIRS{} densely aligned frame pairs.
The dataset is obtained with a semi-supervised strategy by combining self-supervision with sparse annotations to obtain dense inter-frame correspondences.
We also provide a test set along with several metrics for evaluating non-rigid matching and non-rigid reconstruction.
We believe that our dataset is a first step towards enabling learning-based non-rigid matching and our benchmark will help to quantitatively and objectively compare different approaches.

%% file: appendix.tex
\section*{Appendix}

In this appendix, we provide further details about our approach.
Online benchmark is presented in Sec.~\ref{sec:online_benchmark}, network architecture and training is described in Sec.~\ref{sec:network_details}, the details of our least-squares GPU solver are given in Sec.~\ref{sec:solver}, dataset statistics are provided in Sec.~\ref{sec:dataset_statistics}, the dense alignment that is used for training data generation is detailed in Sec.~\ref{sec:dense_alignment_details}, and more qualitative matching and reconstruction results are given in Sec.~\ref{sec:qualitative_results}, including a comparison to KillingFusion~\cite{slavcheva2017killingfusion}.

\section{Online Benchmark}
\label{sec:online_benchmark}
We made dataset publicly available and can be downloaded from \url{https://github.com/AljazBozic/DeepDeform}. We provide an online benchmark for two tasks: optical flow estimation and non-rigid reconstruction. Benchmark is available at \url{http://kaldir.vc.in.tum.de/deepdeform_benchmark}. The evaluation metrics follow the metrics described in the paper. For optical flow we evaluated average pixel error and pixel accuracy (i.e., ratio of pixels with less than $20$ pixel error). For non-rigid reconstruction we evaluate average deformation and geometry error in $cm$.

\section{Network Details}
\label{sec:network_details}
In this section, both the network architecture and the training process are described in detail.
\subsection{Architecture Details}
\begin{figure}[h!]
    \begin{center}
    \includegraphics[width=0.4\columnwidth]{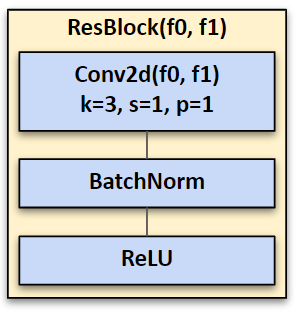}
    \caption{\textbf{Residual block:} the core unit in our network.}
    \label{fig:architecture_resblock}
    \end{center}
\end{figure}

\begin{figure}[h!]
    \begin{center}
    \includegraphics[width=0.9\columnwidth]{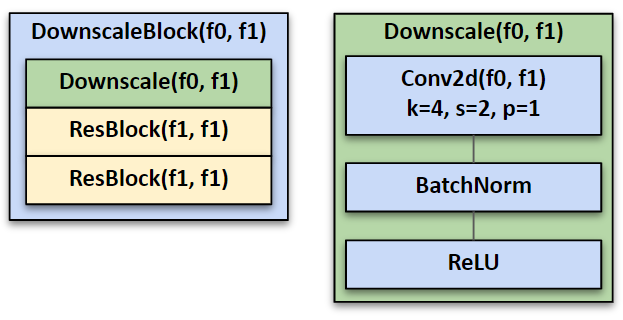}
    \caption{\textbf{Downscale block:} building block of the encoder network; it reduces the input dimension and increases the feature size.}
    \label{fig:architecture_downscale}
    \end{center}
\end{figure}

\begin{figure}[bp!]
    \begin{center}
    \includegraphics[width=0.4\columnwidth]{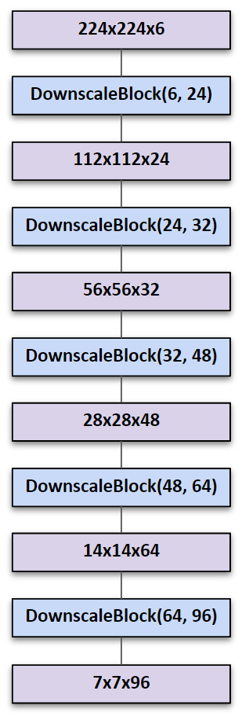}
    \caption{\textbf{Network encoder:} our encoder network takes an RGB-D frame (3 channels for RGB and 3 channels for the backprojected depth points) as input and outputs a $7 \times 7 \times 96$ feature tensor.}
    \label{fig:architecture_encoder}
    \end{center}
\end{figure}

\begin{figure}[bp!]
    \begin{center}
    \includegraphics[width=0.85\columnwidth]{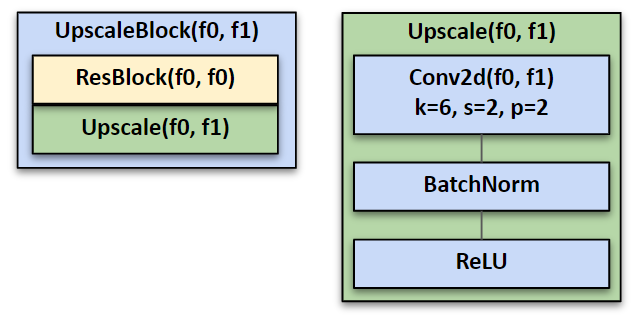}
    \caption{\textbf{Upscale block:} building block of the decoder network; it increases the input dimension and decreases the feature size.}
    \label{fig:architecture_upscale}
    \end{center}
\end{figure}

\begin{figure}[bp!]
    \begin{center}
    \includegraphics[width=\columnwidth]{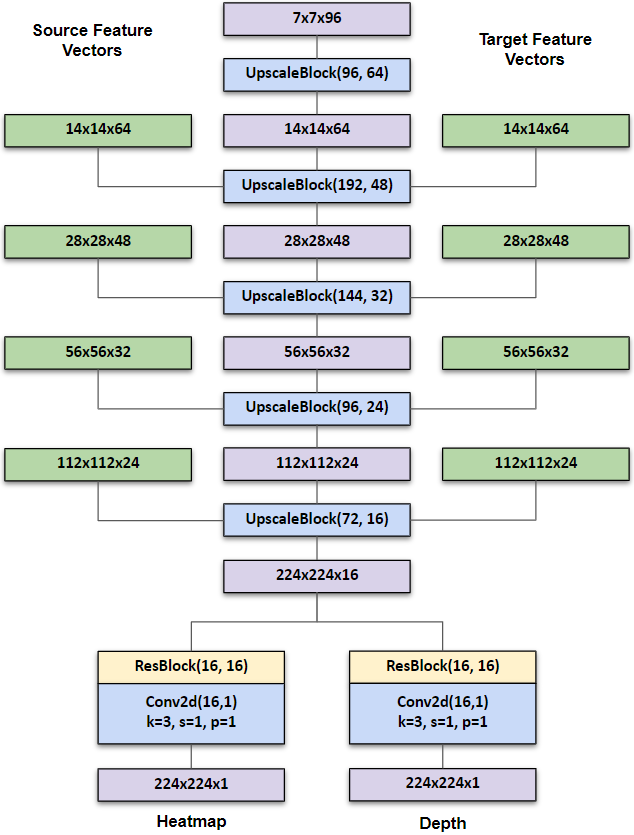}
    \caption{\textbf{Network decoder:} the network decoder takes a feature tensor of dimension $7 \times 7 \times 96$ as input and outputs a heatmap of dimension $224 \times 224 \times 1$ and a depth prediction of dimension $224 \times 224 \times 1$. We employ skip connections that directly provide encoded features of the source and target RGB-D frames to the upscale blocks.} 
    \label{fig:architecture_decoder}
    \end{center}
\end{figure}

\begin{figure}[bp!]
    \begin{center}
    \includegraphics[width=0.45\columnwidth]{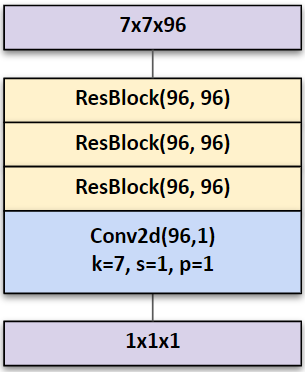}
    \caption{\textbf{Visiblity block:} converts bottleneck features into a visibility score $\in [0, 1]$ that measures whether the source point is visible (high value) or occluded (low value) in the target frame.}
    \label{fig:architecture_visibility}
    \end{center}
\end{figure}

\begin{figure}[bp!]
    \begin{center}
    \includegraphics[width=0.8\columnwidth]{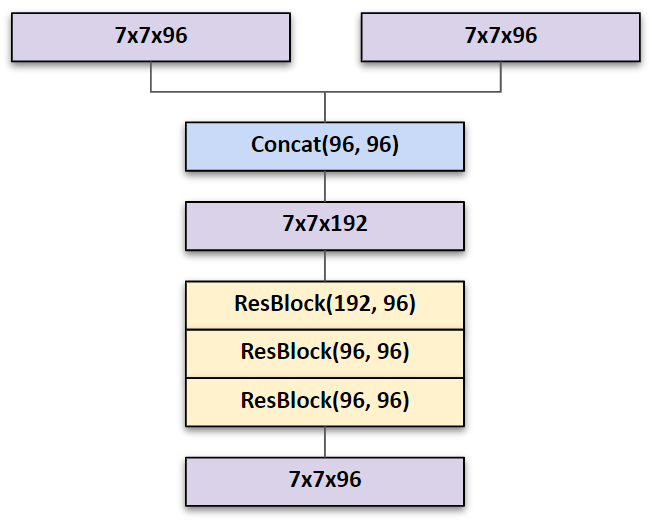}
    \caption{\textbf{Bottleneck:} it combines the feature tensors that correspond to the source and target RGB-D frames.}
    \label{fig:architecture_bottleneck}
    \end{center}
\end{figure}

The network takes as input an RGB-D frame of dimension $224 \times 224 \times 6$, with $3$ channels for RGB and $3$ channels for backprojected depth points.
The network outputs are a heatmap of size $224 \times 224 \times 1$, another heatmap of size $224 \times 224 \times 1$ and a visibility score of size $1 \times 1 \times 1$.
The core unit used in our network is a residual block (as visualized in Fig.~\ref{fig:architecture_resblock}), the other two building blocks are the downscale (see Fig.~\ref{fig:architecture_downscale}) and the upscale (see Fig.~\ref{fig:architecture_upscale}) blocks.
Our Siamese network is based on two towers that share the encoder.
The encoder consists of residual blocks to extract a feature tensor of dimension $7\times7\times96$, with all layers detailed in Fig.~\ref{fig:architecture_encoder}.
Encoded features of the source and target patch are concatenated along the dimension of the channels.
Afterwards, a bottleneck layer (dimension of $7\times7\times96$) is applied to reduce the feature dimension back to $96$, see Fig.~\ref{fig:architecture_bottleneck}.
The network has one decoder that maps the features after the bottleneck layer to a probability heatmap $\mathcal{H}$ and a dense depth prediction $\mathcal{D}$, its layer dimensions are provided in Fig.~\ref{fig:architecture_decoder}.
Our network architecture has skip connections between the encoder and decoder, similar to a U-Net~\cite{ronneberger2015unet}.
In addition to the decoder, there is also a small convolutional network converting bottleneck features of dimension $7\times7\times96$ to a visibility score $\mathcal{O}$ of dimension $1 \times 1 \times 1$, presented in Fig.~\ref{fig:architecture_visibility}.


\subsection{Training Details}
We implemented our non-rigid matching approach in PyTorch \cite{paszke2017automatic} and trained it using stochastic gradient descent with momentum ($m=0.9$) and learning rate $0.01$.
For regularization, we use a weight decay of $0.0005$. 
We use a batch size of $32$.
We divide the learning rate by $10$ every $30$k iteration steps.
We first train the network for heatmap and depth prediction for $100,000$ iterations. 
Afterwards, we train only the visibility detection layers for another $100,000$ iterations, keeping the weights in the encoder and bottleneck layers fixed.
We use different data augmentation techniques, such as random 2D rotation, translation, and horizontal flip.
Every training sample is augmented on-the-fly.

\section{Least Squares GPU Solver}
\label{sec:solver}
We solve the following non-linear energy minimization problem based on a data-parallel Gauss-Newton solver:
\begin{equation}
   \mathcal{G}^* = \argmin_{\mathcal{G}}{E_\textrm{total}(\mathcal{G})} \enspace{.}
\end{equation}
To this end, we reformulate the energy function $E_\textrm{total}$ in terms of a vector field $\mathbf{F}(\mathcal{G})$ by stacking all residuals:
$$E_\textrm{total}(\mathcal{G})=\big|\big|\mathbf{F}(\mathcal{G})\big|\big|_2^2 \enspace{.}$$
In the following, we will drop the dependence on the parameters $\mathcal{G}$ to simplify the notation.
We perform $10$ non-linear Gauss-Newton optimization steps.
In each non-linear optimization step, the vector field $\mathbf{F}$ is first linearized using a first order Taylor expansion.
The resulting linear system of normal equations is then solved based on a data-parallel preconditioned conjugate gradient (PCG) solver.
The normal equations are defined as follows:
$$\mathbf{J}^T\mathbf{J} \boldsymbol\delta = -\mathbf{J}^T\mathbf{F} \enspace{.}$$
Here, $\mathbf{J}$ is the Jacobian matrix of $\mathbf{F}$.
After solving the normal equations the unknowns are updated based on $\boldsymbol\delta$:
$$\mathcal{G}_k = \mathcal{G}_{k-1} + \boldsymbol\delta \enspace{.} $$
Each normal equation is solved based on $20$ iteration steps.
In each PCG step, we first materialize $\mathbf{J}$ in global GPU memory.
The central operation in the PCG solver is to apply the system matrix $\mathbf{J}^T\mathbf{J}$ to the current decent direction.
In order to run in a data-parallel manner, we launch one thread per residual term in $\mathbf{F}$.
Each thread reads the required entries of $\mathbf{J}$ (and therefore also of $\mathbf{J}^T$) and computes its contribution to the partial derivatives of the unknowns.
All contributions are summed up for each unknown using atomic operations.

\section{Dataset Statistics}
\label{sec:dataset_statistics}
\begin{figure}[bp!]
    \includegraphics[width=\columnwidth]{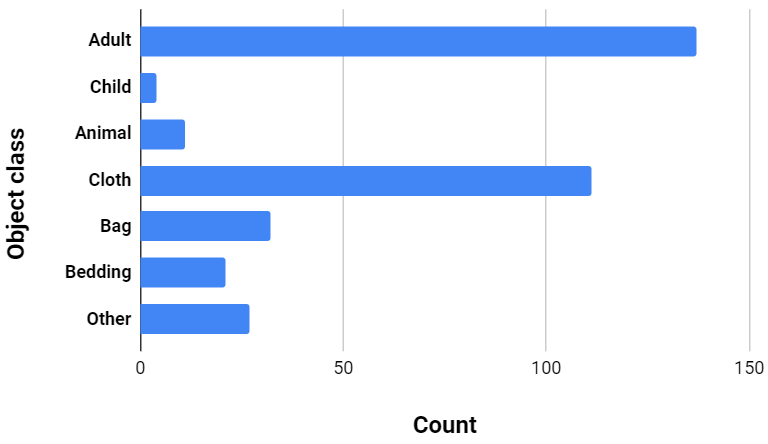}
    \caption{Object class variety: we include many different sequences of dynamic objects, such as cloths, bags, etc.}
    \label{fig:dataset_statistics}
\end{figure}

We provide a train-val-test split using the following distribution of sequences:
340 sequences are in the training set, 30 in the test set, and 30 in the validation set.
We made sure that there is no overlap between captured environments between training and validation/test scenes.
We crowd-sourced a large number of annotations for the recorded RGB-D sequences, which makes our dataset suitable for supervised learning of correspondence matching, see Tab.~\ref{tab:dataset_annotations}.
Our novel dataset for learning non-rigid matching covers a diverse range of non-rigid object classes, as shown in Fig.~\ref{fig:dataset_statistics}, and also includes challenging deformations and camera motion, see Tab.~\ref{tab:dataset_motion}.

\begin{table}
\begin{center}
\begin{tabular}{|l|r|}
\hline
\textbf{Data type} & \textbf{Count} \\ \hline
Object masks & $\MASKS{}$  \\ \hline
Sparse matches & $\MATCHES{}$  \\ \hline
Point occlusions & $\OCCLUSIONS{}$  \\ \hline
Frame pairs & $\PAIRS{}$  \\ \hline
\end{tabular} \\
\caption{\label{tab:dataset_annotations} Dataset annotation statistics, presenting the total number of annotations.}
\end{center}
\end{table}

\begin{table}
\begin{center}
\begin{tabular}{|l|r|}
\hline
\textbf{Motion type} & \textbf{Average motion} \\ \hline
2D change (pixel) & $65.4$  \\ \hline
3D point motion (m) & $0.22$  \\ \hline
\end{tabular} \\
\caption{\label{tab:dataset_motion} Motion and deformation statistics, computed from correspondence annotations. Rigid camera motion is also contained in the motion statistics.}
\end{center}
\end{table}

\section{Dense Alignment Details}
\label{sec:dense_alignment_details}
\begin{figure*}
\includegraphics[width=\linewidth]{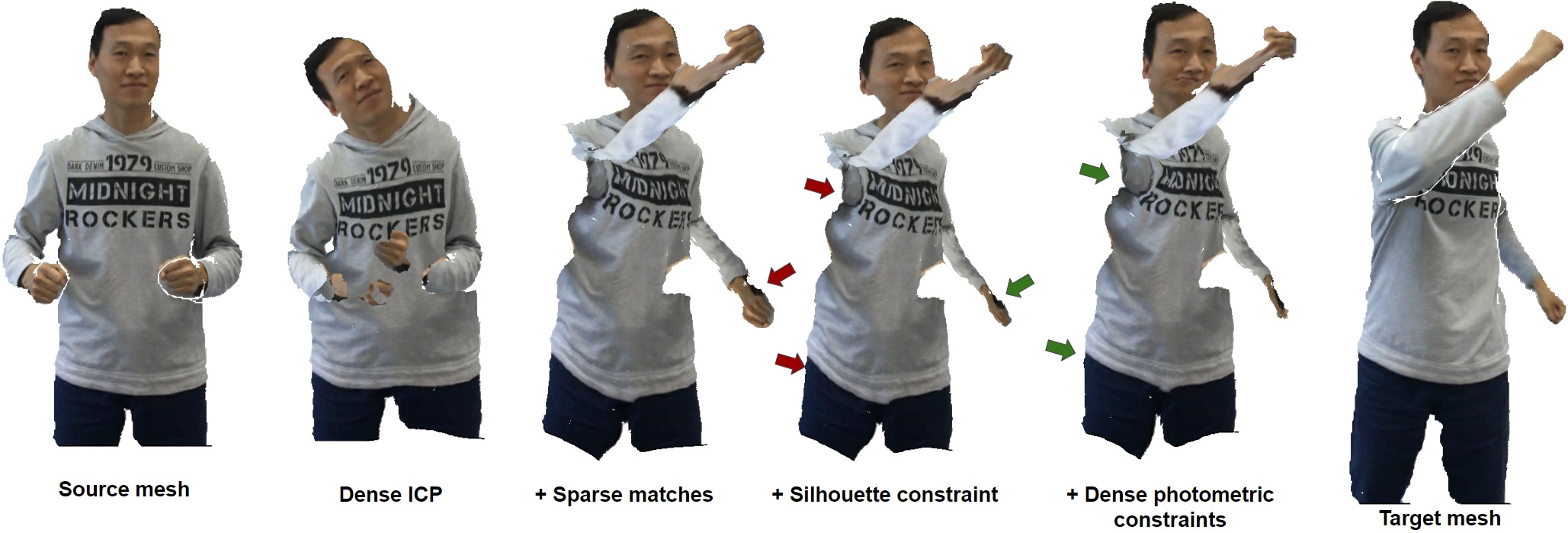} \\
\vspace{-0.5cm}
\caption{
For dense alignment, different optimization constraints are used to align the source mesh to the target mesh. Here we qualitatively show the effect of different constraints on the alignment of the given RGB-D frame pair.
}
\label{fig:dense-alignment-constraints}
\end{figure*}
\begin{figure}
    \centering
    \includegraphics[width=\columnwidth]{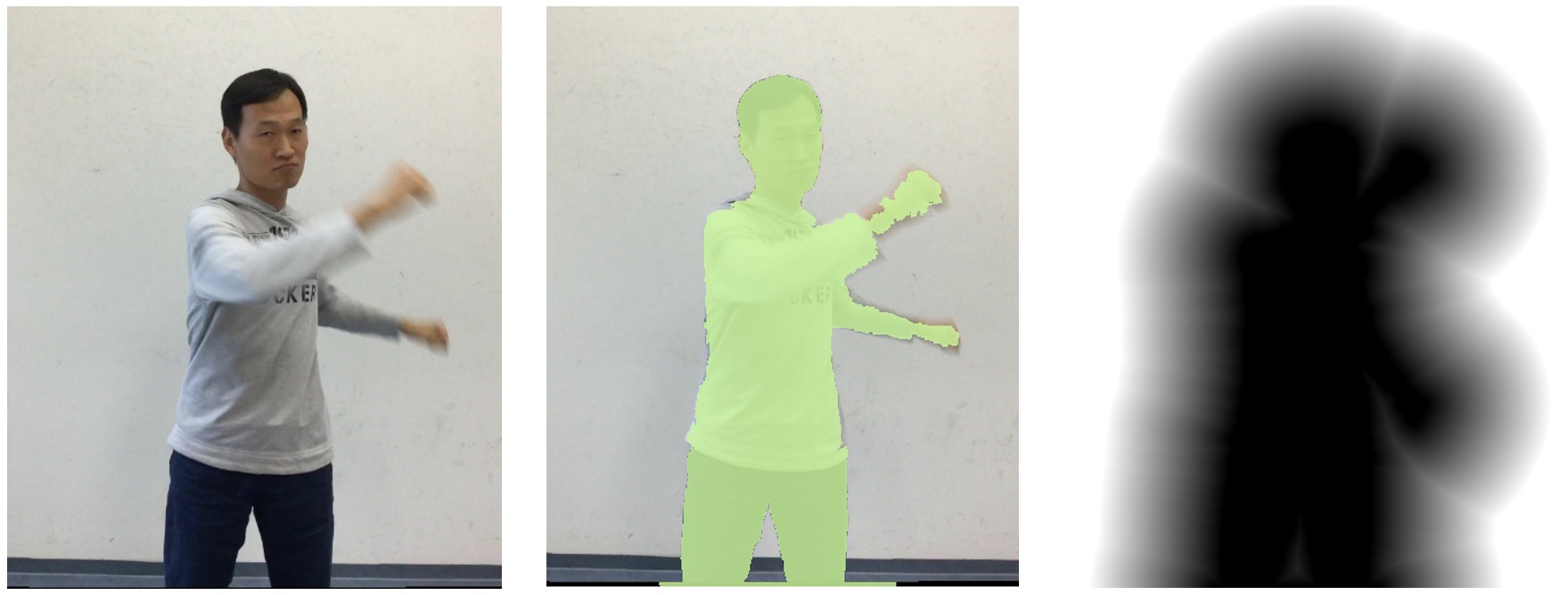}
    \caption{
        For a target object (left), given its mask (middle), we can compute the pixel distance map (right), with zero distances inside the object mask and increasing distances outside the object mask.
    }
    \label{fig:silhouette-map}
\end{figure}

Given two RGB-D frames with dynamic object segmentation masks and sparse correspondences, dense alignment computes dense matches between the source and target RGB-D frames.
Initially, a mesh is extracted from the source frame by back-projecting the pixels into 3D and using pixel-wise triangle connectivity.
All vertices that are outside the source object mask are filtered out.
Afterwards, a deformation graph is sampled uniformly over the source object mesh, and the deformations $\mathcal{T}$ of all nodes that deform the source mesh onto the target mesh are estimated by minimizing the following optimization energy:
\begin{align*}
E_\textrm{total}(\mathcal{T}) = &E_\textrm{data}(\mathcal{T}) + \lambda_\textrm{photo} E_\textrm{photo}(\mathcal{T}) + \lambda_\textrm{silh} E_\textrm{silh}(\mathcal{T})+ \\
&\lambda_\textrm{sparse} E_\textrm{sparse}(\mathcal{T}) + \lambda_\textrm{reg} E_\textrm{reg}(\mathcal{T}) \enspace{.}
\end{align*}
Data term $E_\textrm{data}(\mathcal{T})$ and regularization term $E_\textrm{reg}(\mathcal{T})$ are defined the same as in the traditional non-rigid reconstruction pipeline, minimizing point-to-point and point-to-plane distances, and using as-rigid-as-possible (ARAP) regularization.
As can be seen in Fig.~\ref{fig:dense-alignment-constraints}, dense ICP constraints alone can result in poor frame-to-frame alignment.
A big improvement is achieved by using sparse match constraints, defined as:
\begin{equation*}
    E_{\textrm{sparse}}(\mathcal{T}) = \sum_{(\mathbf{s}_i, \mathbf{t}_i) \in \mathcal{M}}{  \big( \mathcal{W}_\mathcal{T} (\mathbf{s}_i) - \mathbf{t}_i \big)^2 } \enspace{.}
\end{equation*}
Here $\mathbf{s}_i$ and $\mathbf{t}_i$ are annotated match points, and $\mathcal{W}_\mathcal{T} (\bullet)$ is the deformation operator that for a given point takes the nearest deformation nodes and executes linear blending of their deformations.
An additional silhouette constraint encourages all source mesh vertices to be projected inside the target object's mask:
\begin{equation*}
    E_{\textrm{silh}}(\mathcal{T}) = \sum_{\mathbf{v} \in \mathcal{S}}{  \big( \textrm{PixDist} (\Pi (\mathcal{W}_\mathcal{T} (\mathbf{v}))))^2 } \enspace{.}
\end{equation*}
The constraint is computed for every vertex $\mathbf{v}$ in the source mesh, where $\Pi(\bullet)$ is a projection from $3$D to $2$D image space, and $\textrm{PixDist}$ is the pixel distance map, computed as shown in Fig.~\ref{fig:silhouette-map}.
Lastly, dense color constraints enable alignment correction in cases where we have useful texture information:
\begin{equation*}
    E_{\textrm{photo}}(\mathcal{T}) = \sum_{\mathbf{v} \in \mathcal{S}}{  \big( \nabla \textrm{I}_t (\Pi (\mathcal{W}_\mathcal{T} (\mathbf{v}))) - \nabla \textrm{I}_s (\Pi (\mathbf{v})))^2 } \enspace{.}
\end{equation*}
Here $\textrm{I}_s$ and $\textrm{I}_t$ are source and target color images. We use color gradients instead of raw colors to be invariant to constant intensity change.

\section{Qualitative Results}
\label{sec:qualitative_results}
\begin{figure*}
\centering
\includegraphics[width=0.95\linewidth]{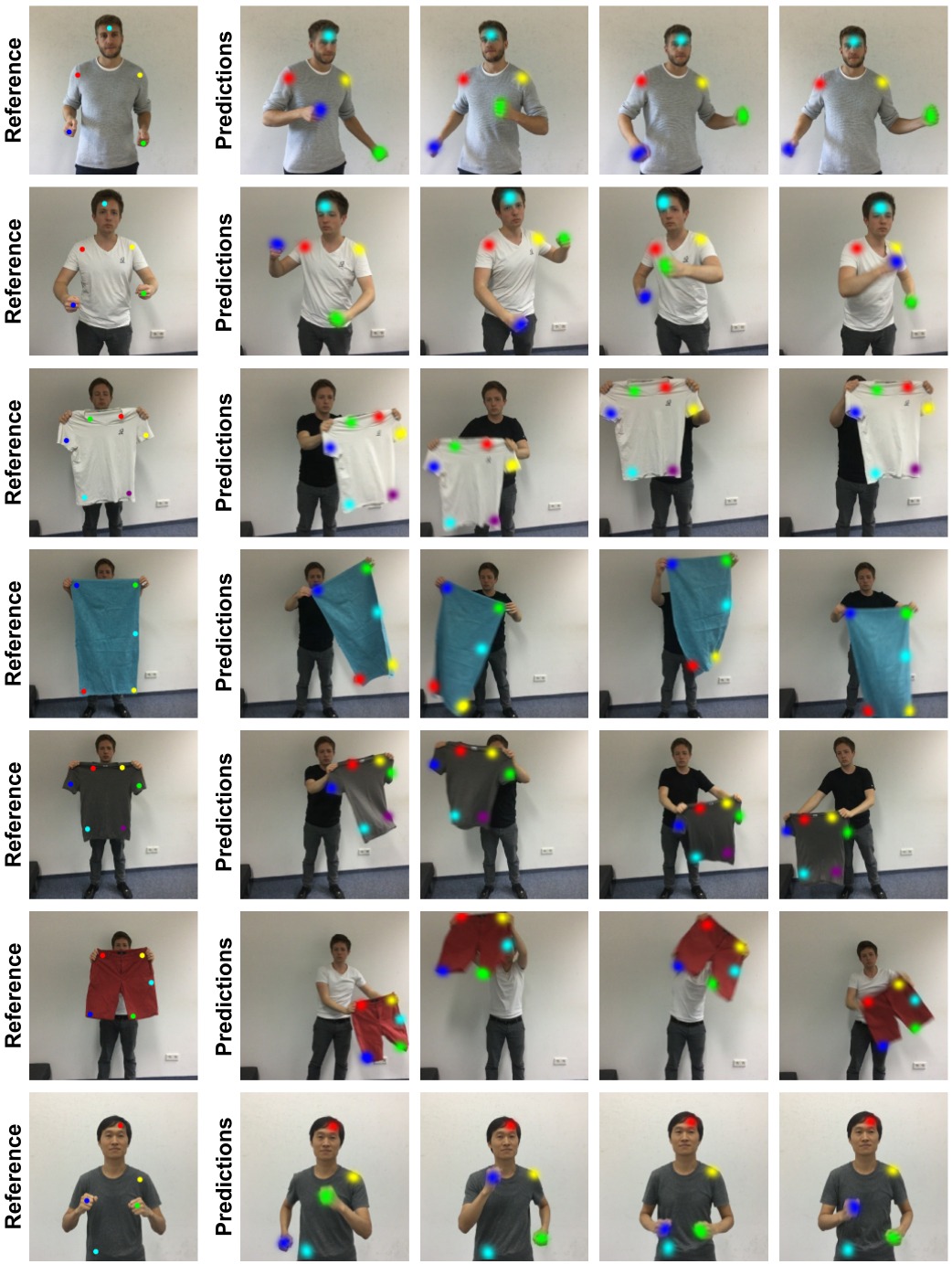} \\
\caption{
Qualitative heatmap prediction results. Our approach works well even for highly non-rigid motions.
}
\label{fig:heatmap_predictions}
\end{figure*}

\begin{figure*}
\centering
\includegraphics[width=0.6\linewidth]{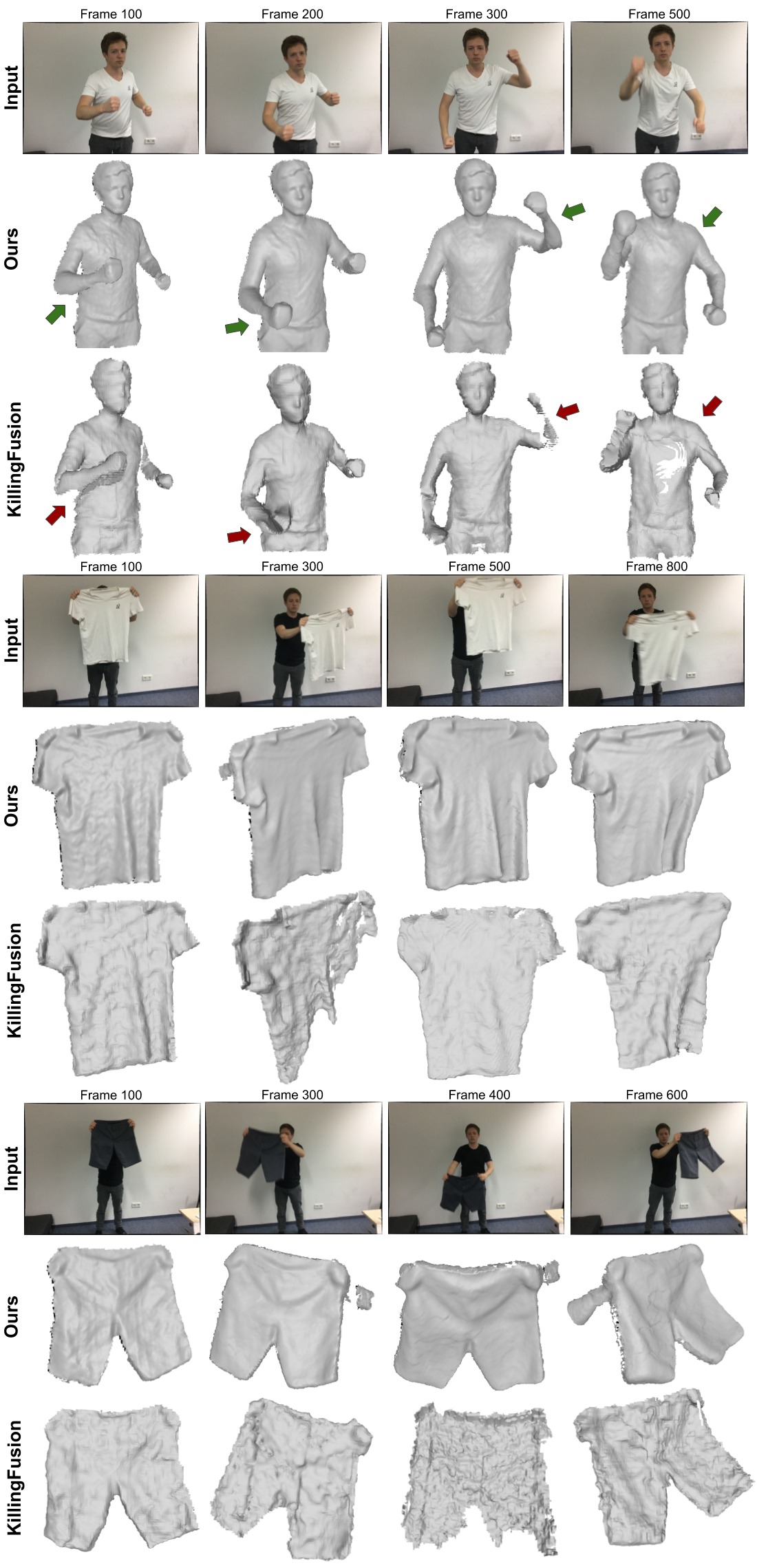} \\
\caption{
Qualitative reconstruction comparison of our approach to KillingFusion~\cite{slavcheva2017killingfusion}. Reconstruction results were kindly provided by the authors. Our approach obtains more robust and higher quality results.
}
\label{fig:qualitative_comparison_killingfusion}
\end{figure*}

\begin{figure*}
\centering
\includegraphics[width=0.6\linewidth]{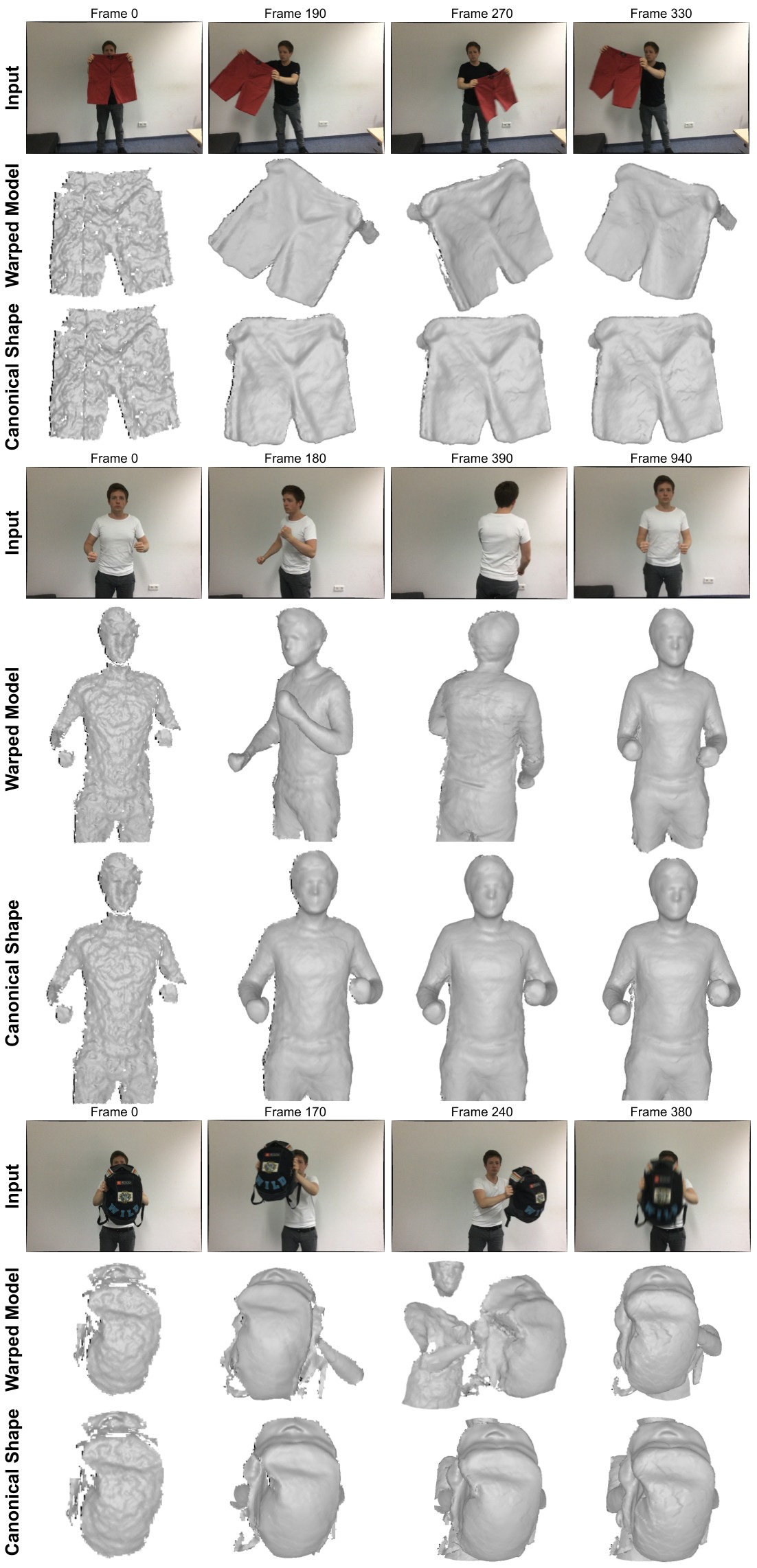} \\
\caption{
Qualitative reconstruction results, showing both the warped model (in the current frame) and the canonical shape (in the reference frame).
Our approach obtains high quality reconstruction results.
}
\label{fig:qualitative_reconstruction}
\end{figure*}

In the following figures we present more qualitative results. More specifically, in Fig.~\ref{fig:heatmap_predictions}, our network heatmap predictions are presented for a set of chosen points, in Fig.~\ref{fig:qualitative_comparison_killingfusion} we provide a qualitative reconstruction comparison to the approach of \cite{slavcheva2017killingfusion}, and in Fig.~\ref{fig:qualitative_reconstruction} reconstruction results of our method are shown, including the shape in the canonical space.
